\documentclass{article}
\usepackage{spconf,amsmath,graphicx,hyperref}
\usepackage{booktabs} 
\usepackage{multirow}
\usepackage{cite}
\usepackage{etoolbox}
\makeatletter
\patchcmd{\thebibliography}
  {\usecounter{enumi}}
  {\setlength{\itemsep}{-0.4em}\usecounter{enumi}}  % -0.4em 可根据需要调节
  {}{}
\makeatother
\usepackage{enumitem}
\usepackage[table]{xcolor}    % 支持表格着色
\usepackage{colortbl}         % 扩展表格颜色
\usepackage{pgf}
\definecolor{lightred}{RGB}{255,230,230}
\definecolor{deepred}{RGB}{76,0,0}
\definecolor{lightblue}{RGB}{230,240,255}
\definecolor{deepblue}{RGB}{0,0,76}

\newcommand{\ACCcell}[2]{%
  \begingroup
  \pgfmathparse{int(#1*100)}% 把 0.8 → 80
  \edef\tempcolor{\noexpand\cellcolor{lightred!\pgfmathresult!deepred}}%
  \tempcolor #2%
  \endgroup
}

\newcommand{\ASAcell}[2]{%
  \begingroup
  \pgfmathparse{int(#1*100)}% 把 0.8 → 80
  \edef\tempcolor{\noexpand\cellcolor{lightblue!\pgfmathresult!deepblue}}%
  \tempcolor #2%
  \endgroup
}
\newcommand{\ACCbest}[1]{\cellcolor{lightred!90!deepred}{{#1}}}
\newcommand{\ASAbest}[1]{\cellcolor{lightblue!90!deepblue}{{#1}}}

\usepackage{url}
\usepackage{arydshln}
\title{Multi-Physics: A Comprehensive Benchmark for Multimodal LLMs Reasoning on Chinese Multi-subject Physics Problems}
\name{Zhongze Luo, Zhenshuai Yin, Yongxin Guo, Zhichao Wang, Jionghao Zhu, Xiaoying Tang$^*$\thanks{$^*$Corresponding author: Xiaoying Tang.}}
\address{School of Science and Engineering, The Chinese University of Hong Kong, Shenzhen, China}
\begin{document}
%\ninept
%
\maketitle
\begin{abstract}
While multimodal LLMs (MLLMs) demonstrate remarkable reasoning progress, their application in specialized scientific domains like physics reveals significant gaps in current evaluation benchmarks. Specifically, existing benchmarks often lack fine-grained subject coverage, neglect the step-by-step reasoning process, and are predominantly English-centric, failing to systematically evaluate the role of visual information. Therefore, we introduce \textbf {Multi-Physics} for Chinese physics reasoning, a comprehensive benchmark that includes 5 difficulty levels, featuring 1,412 image-associated, multiple-choice questions spanning 11 high-school physics subjects. We employ a dual evaluation framework to evaluate 20 different MLLMs, analyzing both final answer accuracy and the step-by-step integrity of their chain-of-thought. Furthermore, we systematically study the impact of difficulty level and visual information by comparing the model performance before and after changing the input mode. Our work provides not only a fine-grained resource for the community but also offers a robust methodology for dissecting the multimodal reasoning process of state-of-the-art MLLMs, and our dataset and code have been open-sourced \footnote{\url{https://github.com/luozhongze/Multi-Physics}}.

\end{abstract}
\section{Introduction}
\label{sec:intro}

With the rapid development of large language models (LLMs), an increasing number of models have demonstrated remarkable performance in logical reasoning and comprehending human knowledge \cite{pang2025language,huang2025improving}. This progress can be attributed to continuous advancements in the knowledge reasoning capabilities of LLMs \cite{yu2024natural}, exemplified by research on step-by-step problem-solving by chain-of-thought (CoT) \cite{xie2025leveraging,kang2025enhancing}, as well as the gradual enrichment of evaluation frameworks \cite{fei2025path}. Concurrently, numerous multimodal large language models (MLLMs) have also exhibited considerable potential \cite{young2024yi,hurst2024gpt,claude-3.5-sonnet,team2025kimi,gpt-4.1,claude-4,bai2023qwen,QVQ-Max,o4-mini,comanici2025gemini,Mistral,team2025gemma,zhu2025internvl3,Meta,qwen2.5,hong2025glm}. Notably, these models can additionally incorporate visual understanding \cite{guotrace,wang2025cof}, enabling logical reasoning across a broader spectrum of visual knowledge tasks \cite{yang2024empowering,guo2025vtg}.

Nowadays, there have been many studies on multi-task evaluation \cite{tseng2024av}, especially in the field of mathematical problems \cite{liao2025look,li2025step,zhang2025diagram}. However, physics encompasses a wealth of theorems and constraints, and physics problems across different domains may exhibit distinct visual characteristics. Some existing physical benchmarks are merely single modalities of text, like TPBench\cite{chung2025theoretical}, PHYBench\cite{qiu2025phybench}, ABench-Physics\cite{zhang2025abench}, and UGPhysics\cite{xu2025ugphysics}, and other multimodal physical benchmarks still lack sufficient coverage of fine-grained physics subjects, like PHYSICS\cite{xu2025ugphysics}, PhysReason\cite{zhang2025physreason}, and SEEPHYS\cite{xiang2025seephys}. Furthermore, most current benchmarks often neglect the step-by-step evaluation of the chain of thought and fail to adequately evaluate the impact of visual information (VI) on reasoning \cite{zhang2024mathverse}. Therefore, developing a comprehensive benchmark for assessing the physical knowledge reasoning capabilities of MLLMs, integrated with CoT-based progressive evaluation and VI comparative evaluation holds significant academic value.

We present a comprehensive multimodal benchmark covering diverse high-school physics fields: {\bf Multi-Physics}. It consists of 11 specialized subjects and 1,412 problems, each presented as an image-associated multiple-choice question, and all questions are assessed on 5 difficulty levels. Unlike most existing physics benchmarks that primarily focus on English, it emphasizes Chinese comprehension, enabling a more robust evaluation of MLLMs' understanding of Chinese-based physics problems. To explore the extent to which VI influences the problem-solving reasoning of different MLLMs, we also set the problem image as a variable during the evaluation process and conduct a CoT evaluation without introducing the problem image as well. The key contributions of our work are as follows:

\begin{itemize}[nosep, leftmargin=*]
\item We develop and introduce Multi-Physics to evaluate 20 different MLLMs on Multi-subject Chinese physics problems, which features 1,412 image-associated, multiple-choice questions across 11 subjects and 5 difficulty levels.
\item We propose a dual-pronged evaluation method framework, which combines the evaluation of answer accuracy with the step-by-step progressive evaluation of the model's CoT.
\item We conduct a systematic investigation into the impact of difficulty level and VI on MLLMs reasoning, which provides crucial empirical data for understanding the multimodal reasoning mechanisms in state-of-the-art MLLMs.
\end{itemize}

\section{Methods}
\label{sec:Methods}

\subsection{Multi-Physics Construction}
\label{sec:construction}
Due to the lack of Chinese multimodal physics benchmarks, we collect and open-source the Multi-Physics, constructed with three key stages: data collection, data filtering and standardization, and data annotation.

\textbf{Data collection.} We collect over 2,000 real exam papers in PDF format from publicly available practice questions, simulation tests, and physics competitions, covering physics examination questions across all three years of Chinese high school curricula. We use the Mathpix API\footnote{\url{http://mathpix.com/}} to perform optical character recognition (OCR) on the PDFs, converting them into markdown text and extracting associated question images. Subsequently, annotators familiar with high school physics knowledge and qualified through standardized exams manually verify and correct the grammar and \LaTeX{} formatting of all questions using SimpleTex\footnote{\url{https://www.simpletex.net/}}.

\textbf{Data filtering and standardization.} We filter the questions in the markdown text, retaining only the multiple-choice questions containing the question images and accompanying solution analysis, and remove duplicate or non-compliant questions before converting them into JSON format with a fixed four-tuple structure, an example is shown in Fig. \ref{fig:example}.

\begin{figure}[!h]
    \centering
    \includegraphics[width=0.9\linewidth]{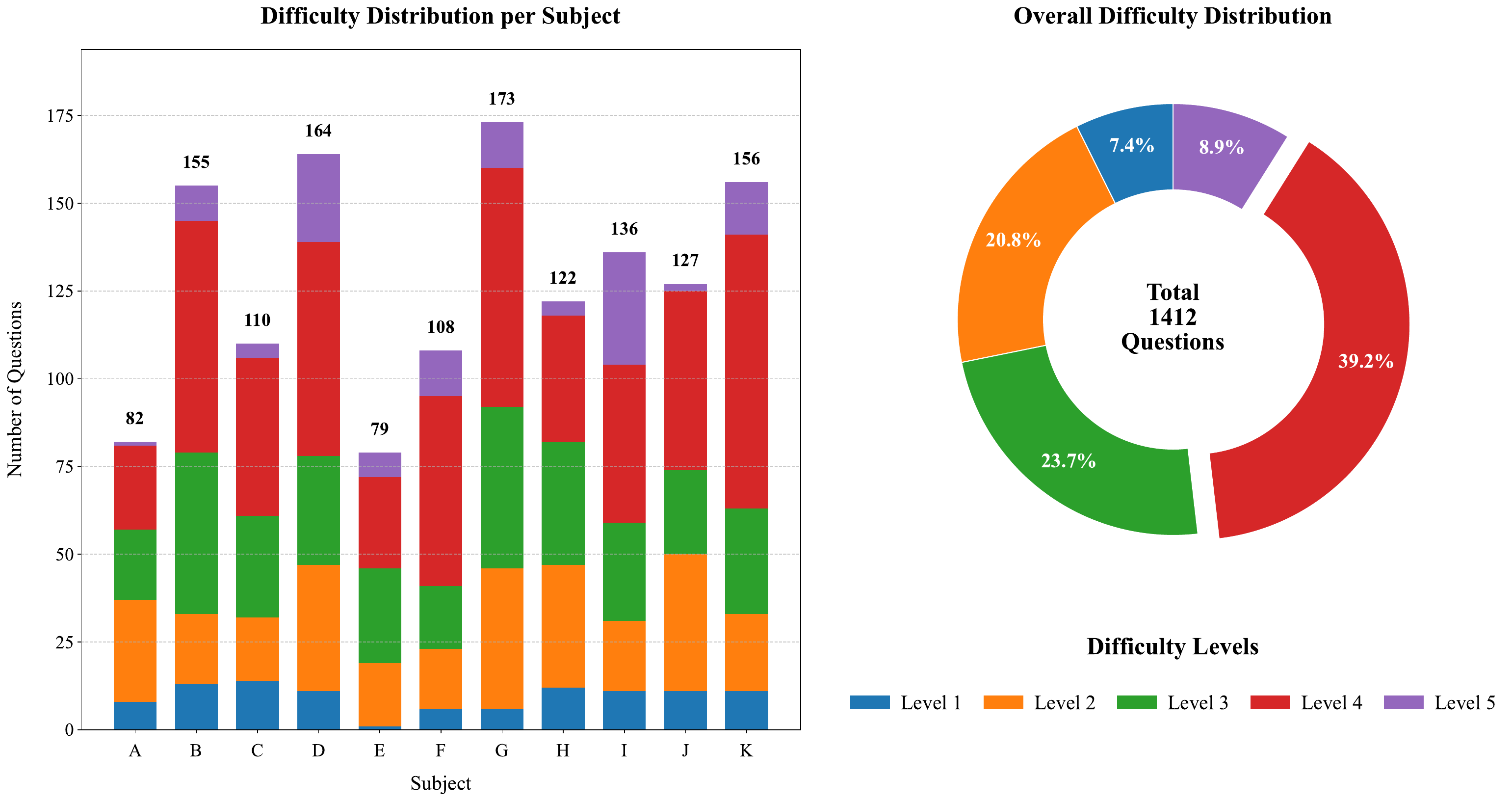}
    \caption{Distribution of difficulty levels.}
    \label{fig:level}
\end{figure}

\textbf{Data annotation.} We divide the questions into 11 different subjects. For the questions where the data source has already provided labels that match these subjects, we retain them in the matching subject. For other problems, we adopt a two-stage fine-grained classification method to classify them by subject. Firstly, we use GPT-4.1\cite{gpt-4.1} to make subject judgments about them and filter out problems that do not belong to them. Then, we ask annotators with rich knowledge of physics to review and check the existing judgment results. Finally, we divide all the questions into 11 different JSON files according to the classified subjects and number them in letters A-K. Also, for all questions, we reuse GPT-4.1 to assess the difficulty levels from 1 to 5 from the perspective of a ``high school physics competition teacher''. The difficulty level is related to the number of calculation steps, the abstraction of the required knowledge points, or the complexity of the graphic and textual information, ensuring the fairness and professionalism of the assessment. The field name, number of questions, average question length (AQL), and average analysis length (AAL) for subjects are shown in Table \ref{tab:1}. The distribution of difficulty levels is shown in Fig. \ref{fig:level}.

\begin{table}[!ht]
    \scriptsize
    \centering
    \caption{Statistics of subject questions.}
    \begin{tabular}{ccccc}
    \toprule
    \bf Letters & \bf Subject & \bf Numbers  & \bf AQL & \bf AAL  \\ 
    \midrule
    A & Linear Motion & 82 & 168.99 & 238.15 \\ 
    B & Interactions in Mechanics & 155 & 201.05 & 218.69 \\ 
    C & Newton's Laws of Motion & 110 & 201.33 & 234.85 \\ 
    D & Curvilinear Motion & 164 & 203.45 & 237.64 \\ 
    E & Law of Universal Gravitation & 79 & 250.52 & 316.43 \\ 
    F & Mechanical Energy & 108 & 218.66 & 274.52 \\ 
    G & Electrostatic Field & 173 & 207.30 & 212.38 \\ 
    H & Constant Electric Current & 122 & 183.20 & 215.63 \\ 
    I & Magnetic Field & 136 & 245.34 & 262.21 \\ 
    J & Electromagnetic Induction & 127 & 197.37 & 195.15 \\ 
    K & Alternating Current & 156 & 200.90 & 270.12 \\ 
    \midrule
    \bf All & \bf Multi-Physics & \bf 1,412 & \bf 206.75 & \bf 239.74 \\
    \bottomrule
    \end{tabular}
    \label{tab:1}
\end{table}

\textbf{Advantages of Multi-Physics benchmark.} Compared to existing physics reasoning benchmarks, Multi-Physics offers a more comprehensive and diverse evaluation framework. While prior datasets like PHYBench \cite{qiu2025phybench} and ABench-Physics \cite{zhang2025abench} lack visual inputs, and others such as PHYSICS \cite{feng2025physics} or PhysReason \cite{zhang2025physreason} are limited in subject coverage, Multi-Physics bridges these gaps by combining 1,412 questions with 1,438 images across 11 subjects. This balance of multimodal data and wide-ranging physics subjects enables more robust evaluation of models' conceptual understanding and physics problem-solving abilities, as shown in Table \ref{tab:2}.

\begin{table}[!t]
    \scriptsize
    \centering
    \caption{Comparison of Multi-Physics and other benchmarks.}
    \begin{tabular}{lccc}
    \toprule
    \bf Benchmark & \bf Images & \bf Sizes & \bf Subjects \\ 
    \midrule
    TPBench\cite{chung2025theoretical} & 0 & 57 & 4 \\
    PHYBench\cite{qiu2025phybench} & 0 & 500 & 6 \\ 
    ABench-Physics\cite{zhang2025abench} & 0 & 500 & 9 \\  
    UGPhysics\cite{xu2025ugphysics} & 0 & 11,040 & 13  \\ 
    PHYSICS\cite{feng2025physics} & 298 & 1,297 & 6 \\ 
    PhysReason\cite{zhang2025physreason} & 972 & 1,200 & 7 \\ 
    SEEPHYS\cite{xiang2025seephys} & 2,245 & 2,000 & 7 \\ 
    \midrule
    \bf Multi-Physics & \bf 1,438 & \bf 1,412 & \bf 11 \\
    \bottomrule
    \end{tabular}
    \label{tab:2}
\end{table}

\subsection{Evaluation Methods}

We propose a dual-pronged evaluation method framework, which combines the evaluation of answer accuracy with the step-by-step progressive evaluation of the model's CoT.

\textbf{The evaluation of answer accuracy.} For each question $q$ from the total set of questions $Q$, we define a scoring function $S(q)$ that compares the model's answer ($A_m$) with the standard answer ($A_{std}$). The function is defined as $S(q) = 1, \text{ if } A_m = A_{std}; 0.5, \text{ if } A_m \neq A_{std} \text{ and } A_m \subseteq A_{std}; 0, \text{ otherwise}$. This formula first checks for an exact match. If they are not identical, it then proceeds to check if the model's answer is a substring of the standard answer. The overall model accuracy, $\text{ACC}_{\text{total}}$, is the average score across all questions: $\text{ACC}_{\text{total}} = \frac{\sum_{q \in Q} S(q)}{|Q|}$.

\textbf{The evaluation of the model's CoT.} For each question $q$, we first extract the sequence of reasoning steps, denoted as $R(q) = \{r_1, r_2, \dots, r_n\}$, where $n = |R(q)|$ is the total number of steps for that question. Instead of direct string matching, we employ a model-based evaluation strategy. Each individual step $r_i \in R(q)$ is submitted to Gemini-2.5-Flash\cite{comanici2025gemini} to judge its correctness. We define a binary judgment function, $J(r_i)$, which yields 1 if the step $r_i$ is judged as ``correct'' and 0 otherwise, and calculate the step accuracy for a single question $q$, denoted $\text{ACC}_{\text{step}}(q)$, as the proportion of correct steps: $\text{ACC}_{\text{step}}(q) = \frac{\sum_{i=1}^{n} J(r_i)}{n} = \frac{\sum_{r \in R(q)} J(r)}{|R(q)|}$. Finally, the model's overall reasoning ability is quantified by the Average Step Accuracy (ASA), which is the arithmetic mean of the step accuracies across all questions in the total set $Q$: $\text{ASA} = \frac{\sum_{q \in Q} \text{ACC}_{\text{step}}(q)}{|Q|}$. To measure the average length of the model's generative inference, we define the Average Step Count (ASC) of the model as the arithmetic mean of the number of steps of problems in the dataset: $\text{ASC} = \frac{\sum_{q \in Q} |R(q)|}{|Q|}$.

\begin{figure}[!htbp]
    \centering
    \includegraphics[width=0.9\linewidth]{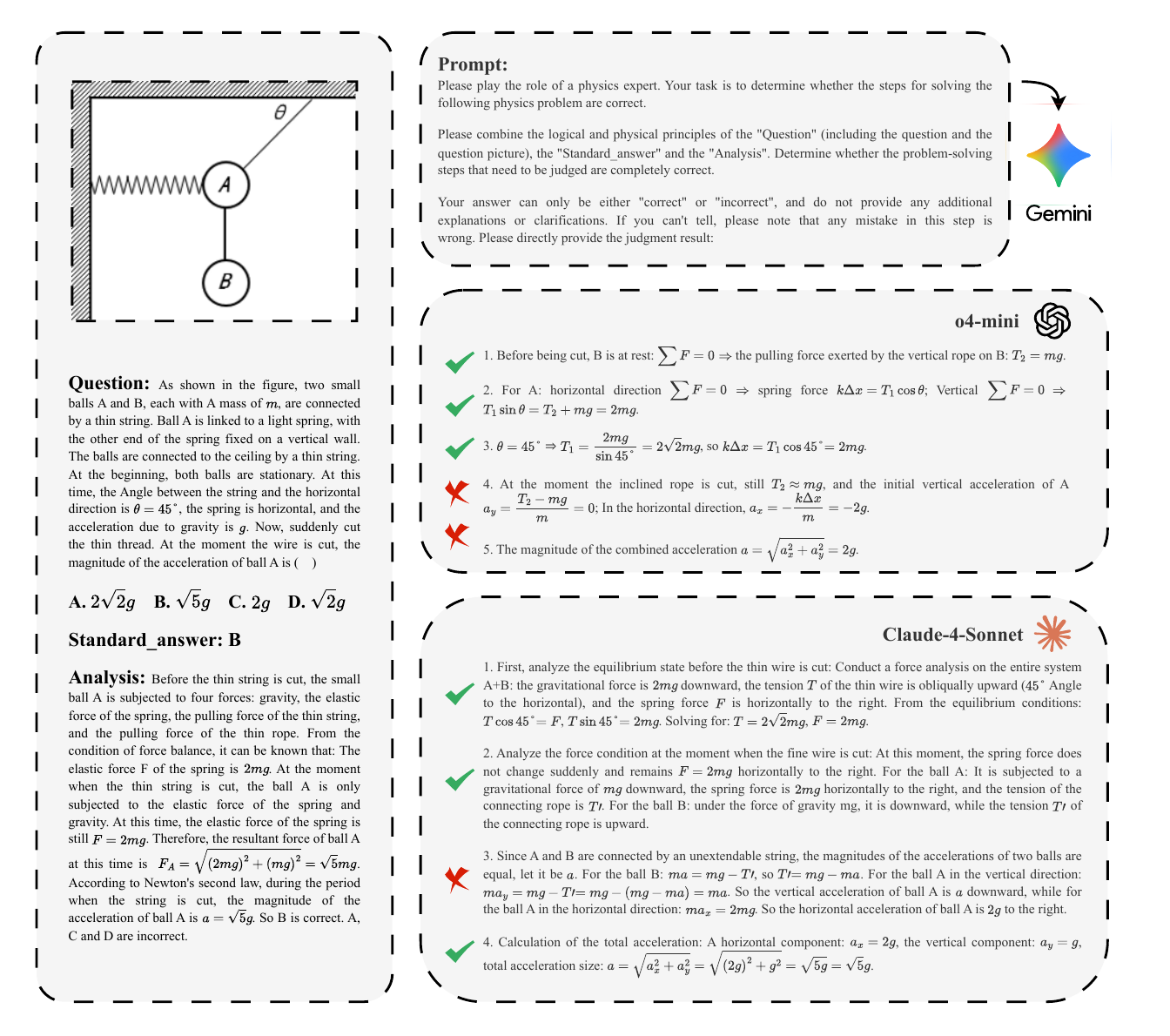}
    \caption{Example of dataset and CoT evaluation. (Translated)}
    \label{fig:example}
\end{figure}

\textbf{Fig. \ref{fig:example} shows an example of our CoT evaluation.} This is the evaluation result of the o4-mini and Claude-4-Sonnet models on the problem in Fig. \ref{fig:example}. Both correctly analyze the initial equilibrium state. However, after that, the former starts from a fundamentally wrong concept (the rope tension does not change suddenly), resulting in all subsequent steps being wrong; the latter wrongly assumes that A and B have the same magnitude of acceleration and makes a mistake about the direction of the vertical force on A, but happens to piece together the correct answer. By evaluating the model's CoT, we can precisely point out the logical flaws in the steps rather than merely judging the correctness of the final answer.

\section{Experiments and Analysis}
\label{sec:Exp}

Our evaluation covers 20 different MLLMs, divided into two main categories: closed-source and open-source models. The closed-source models include: Yi-Vision-V2\cite{young2024yi}, GPT-4o\cite{hurst2024gpt}, Claude-3.5-Sonnet\cite{claude-3.5-sonnet}, Moonshot-V1-128k-Vision-Preview\cite{team2025kimi}, GPT-4.1\cite{gpt-4.1}, ChatGPT-4o-Latest\cite{hurst2024gpt}, Claude-4-Sonnet\cite{claude-4}, Qwen-VL-Max\cite{bai2023qwen}, QvQ-Max\cite{QVQ-Max}, o4-mini\cite{o4-mini}, Gemini-2.5-Flash\cite{comanici2025gemini}, and Gemini-2.5-Pro\cite{comanici2025gemini}. The open-source models include: Mistral-Small-3.2-24B-Instruct\cite{Mistral}, Gemma-3-27B-It\cite{team2025gemma}, InternVL3-14B\cite{zhu2025internvl3}, Llama-4-Scout-17B\cite{Meta}, Qwen2.5-VL-32B-Instruct\cite{qwen2.5}, GLM-4.1V-9B-Thinking\cite{hong2025glm}, Qwen2.5-VL-72B-Instruct\cite{qwen2.5}, and Llama-4-Maverick-17B\cite{Meta}. All models are evaluated under two input modes: with images (w/) and without images (w/o), and we also present the evaluation results of the two methods: Accuracy (ACC) and Average Step Accuracy/Average Step Count (ASA/ASC). The evaluation results on Multi-Physics are shown in Table \ref{tab:complete_restructured}.

\begin{table*}[!ht]
    \tiny 
    \centering
    \caption{The evaluation results of Multi-Physics. (The maximum value of each column in the sub-table is highlighted.)}
    \setlength{\tabcolsep}{3pt}
    \begin{tabular}{lcccccccccccccccccccccccc}
    \toprule
        \multirow{3}{*}{\bf Models} & \multicolumn{12}{c}{\bf ACC} & \multicolumn{12}{c}{\bf ASA/ASC} \\
        \cmidrule(lr){2-13} \cmidrule(lr){14-25}
          & \bf A & \bf B & \bf C & \bf D & \bf E & \bf F & \bf G & \bf H & \bf I & \bf J & \bf K & \bf All & \bf A & \bf B & \bf C & \bf D & \bf E & \bf F & \bf G & \bf H & \bf I & \bf J & \bf K & \bf All \\ 
        \midrule
        \multicolumn{25}{c}{\bf w/ images }\\
        \midrule
        Yi-Vision-V2\cite{young2024yi}  & 14.0 & 11.6 & 17.7 & 20.1 & 17.7 & 20.4 & 11.6 & 23.8 & 20.2 & 17.7 & 11.5 & 16.6 & 24/5.6 & 24/6.5 & 26/5.4 & 31/6.3 & 36/5.1 & 20/5.4 & 10/5.6 & 29/6.4 & 29/6.1 & 32/5.0 & 31/5.4 & 26/5.8 \\ 
        GPT-4o\cite{hurst2024gpt}  & 23.2 & 20.0 & 25.0 & 29.3 & 33.5 & 24.5 & 24.0 & 31.6 & 29.8 & 29.5 & 29.5 & 27.1 & 43/5.4 & 40/5.6 & 44/5.3 & 52/5.3 & 61/4.6 & 42/5.2 & 47/5.3 & 49/5.5 & 45/5.5 & 51/4.9 & 54/5.1 & 48/5.3 \\ 
        Claude-3.5-Sonnet\cite{claude-3.5-sonnet} & 34.8 & 26.8 & 35.0 & 33.8 & 44.9 & 37.0 & 24.9 & 38.9 & 28.3 & 29.1 & 27.6 & 31.8 & 49/4.8 & 43/5.0 & 48/5.1 & 51/5.1 & 69/4.5 & 47/5.1 & 50/4.9 & 51/5.0 & 48/5.0 & 48/4.7 & 56/4.9 & 50/4.9 \\ 
        Moonshot-V1-128k-Vision\cite{team2025kimi} & 26.2 & 23.9 & 33.2 & 31.7 & 51.9 & 42.1 & 28.0 & 39.3 & 29.4 & 29.5 & 35.3 & 32.8 & 38/4.8 & 31/5.4 & 42/5.0 & 44/5.1 & 60/4.7 & 46/5.0 & 40/5.0 & 52/5.3 & 39/5.0 & 47/4.7 & 52/5.1 & 44/5.0 \\ 
        GPT-4.1\cite{gpt-4.1}  & 32.9 & 21.6 & 37.3 & 38.7 & 39.9 & 39.8 & 36.1 & 44.7 & 33.5 & 35.8 & 34.0 & 35.4 & 56/6.0 & 46/7.5 & 54/6.3 & 58/6.6 & 69/5.0 & 50/6.2 & 58/6.0 & 57/6.0 & 49/6.5 & 61/5.6 & 58/6.1 & 56/6.2 \\ 
        ChatGPT-4o-Latest\cite{hurst2024gpt} & 31.1 & 29.0 & 34.1 & 37.2 & 41.1 & 40.3 & 35.3 & 46.3 & 40.4 & 33.1 & 36.2 & 36.5 & 51/5.3 & 46/6.1 & 52/6.1 & 58/6.0 & 70/4.6 & 56/5.9 & 55/5.7 & 57/5.7 & 52/6.1 & 59/5.7 & 54/5.4 & 55/5.7 \\ 
        Claude-4-Sonnet\cite{claude-4} & 39.6 & 45.8 & 51.8 & 57.9 & 66.5 & 55.6 & 41.6 & 52.5 & 41.2 & 48.0 & 52.9 & 49.8 & 44/5.8 & 57/7.3 & 61/6.1 & 72/6.3 & 83/5.6 & 65/5.9 & 62/6.4 & 56/6.2 & 55/6.3 & 63/5.8 & 64/6.1 & 62/6.2 \\ 
        Qwen-VL-Max\cite{bai2023qwen} & 54.9 & 55.2 & 60.0 & 58.5 & 79.1 & 66.7 & 46.2 & 56.1 & 50.0 & 55.5 & 62.8 & 57.5 & 51/5.0 & 50/4.9 & 56/4.8 & 62/4.7 & 82/4.2 & 66/4.6 & 54/4.8 & 64/5.3 & 44/4.8 & 59/4.3 & 64/5.1 & 60/4.8 \\ 
        QvQ-Max\cite{QVQ-Max} & 62.2 & 51.9 & 68.2 & 66.2 & 74.1 & 69.9 & 52.9 & 65.6 & 53.3 & 62.2 & 59.9 & 61.3 & 43/3.4 & 52/4.3 & 69/4.0 & 64/3.7 & 79/2.8 & 68/3.2 & 57/4.0 & 60/3.8 & 50/4.4 & 55/3.6 & 64/4.5 & 58/4.1 \\ 
        o4-mini\cite{o4-mini} & 58.5 & 60.0 & 64.1 & 68.3 & 69.0 & 67.1 & 69.1 & 60.7 & 62.1 & 56.7 & 49.7 & 62.2 & 68/4.5 & 61/4.4 & 68/4.4 & 71/4.5 & 82/4.1 & 71/4.5 & 74/4.3 & 68/4.5 & 60/4.7 & 64/4.4 & 68/4.4 & 68/4.4 \\ 
        Gemini-2.5-Flash\cite{comanici2025gemini}  & \ACCbest{74.4} & 75.2 & 75.5 & 75.9 & 88.0 & 82.4 & 72.0 & 73.0 & 56.6 & 59.1 & 65.1 & 71.6 & 89/4.9 & \ASAbest{81/5.0} & 88/4.7 & 87/5.1 & \ASAbest{96/4.6} & 87/4.8 & 81/4.9 & 82/5.3 & 69/5.2 & 72/4.5 & 82/5.0 & 82/4.9 \\ 
        Gemini-2.5-Pro\cite{comanici2025gemini} & 72.6 & \ACCbest{75.8} & \ACCbest{84.5} & \ACCbest{79.0} & \ACCbest{93.0} & \ACCbest{92.6} & \ACCbest{81.8} & \ACCbest{75.8} & \ACCbest{70.6} & \ACCbest{73.6} & \ACCbest{71.2} & \ACCbest{78.4} & \ASAbest{90/4.6} & 77/5.0 & \ASAbest{91/5.3} & \ASAbest{88/5.2} & \ASAbest{96/5.0} & \ASAbest{90/5.1} & \ASAbest{86/4.7} & \ASAbest{85/5.3} & \ASAbest{73/4.7} & \ASAbest{79/4.6} & \ASAbest{85/5.1} & \ASAbest{85/5.0} \\ 
        \addlinespace[0.5em]
        \hdashline
        \addlinespace[0.5em]
        Mistral-Small-3.2-24B\cite{Mistral}  & 23.2 & 16.5 & 23.2 & 19.8 & 17.7 & 19.9 & 17.3 & 32.4 & 21.7 & 16.9 & 21.8 & 20.7 & 33/5.5 & 31/6.6 & 36/8.7 & 45/6.1 & 54/5.7 & 40/5.8 & 38/5.5 & 51/5.9 & 39/5.6 & 40/5.0 & 42/5.3 & 41/6.0 \\ 
        Gemma-3-27B-It\cite{team2025gemma} & 25.6 & 19.0 & 3.0 & 25.3 & 29.7 & 30.1 & 16.5 & 24.2 & 27.9 & 17.3 & 21.8 & 23.6 & 30/6.2 & 28/8.3 & 32/9.1 & 39/7.6 & 49/6.5 & 35/8.6 & 29/6.9 & 38/7.8 & 28/8.7 & 33/5.9 & 34/6.6 & 34/7.5 \\ 
        InternVL3-14B\cite{zhu2025internvl3}  & 37.2 & 35.2 & 40.9 & 40.2 & 54.4 & 45.8 & 30.3 & 48.0 & 35.7 & 33.5 & 38.5 & 39.0 & 48/5.2 & 38/5.2 & 48/5.2 & 51/5.2 & 59/4.0 & 51/4.7 & 39/4.8 & 56/5.3 & 41/4.9 & 46/4.3 & 51/4.4 & 47/4.9 \\ 
        Llama-4-Scout-17B\cite{Meta} & 31.1 & 35.2 & 41.8 & 43.0 & 53.8 & 51.9 & 36.4 & 42.6 & 36.4 & 33.5 & 43.9 & 40.4 & 38/4.5 & 34/4.6 & 44/4.9 & 49/4.7 & 56/4.0 & 49/4.6 & 36/4.1 & 48/5.0 & 37/4.8 & 41/4.0 & 46/3.9 & 43/4.5 \\ 
        Qwen2.5-VL-32B\cite{qwen2.5} & 42.7 & 45.5 & 49.1 & 43.9 & 60.1 & 56.9 & 36.1 & 49.2 & 32.7 & 40.9 & 50.0 & 45.1 & 40/2.9 & 29/2.9 & 48/3.6 & 44/3.1 & 52/3.1 & 43/3.0 & 40/3.3 & 47/3.6 & 36/3.3 & 41/3.0 & 46/3.7 & 42/3.2 \\ 
        GLM-4.1V-9B-Thinking\cite{hong2025glm} & 47.0 & 42.3 & 57.7 & 56.1 & 69.6 & 59.7 & 51.2 & 57.8 & 35.7 & 37.0 & 53.2 & 50.7 & \ASAbest{60/4.7} & 47/4.9 & \ASAbest{64/4.9} & 64/5.0 & \ASAbest{79/4.3} & 64/4.7 & \ASAbest{57/4.6} & \ASAbest{66/4.9} & 46/4.6 & 53/4.5 & \ASAbest{65/4.8} & \ASAbest{59/4.7} \\ 
        Qwen2.5-VL-72B\cite{qwen2.5} & 43.3 & 42.9 & 57.3 & 59.1 & 67.1 & 66.2 & 44.5 & \ACCbest{57.0} & 41.2 & 46.9 & 58.0 & 52.3 & 51/4.7 & 46/4.9 & 60/4.7 & \ASAbest{65/4.8} & \ASAbest{79/4.4} & 63/4.5 & 47/4.4 & \ASAbest{66/4.8} & 45/4.6 & 56/4.3 & 64/4.7 & 57/4.6 \\ 
        Llama-4-Maverick-17B\cite{Meta} & \ACCbest{51.8} & \ACCbest{60.6} & \ACCbest{66.8} & \ACCbest{69.5} & \ACCbest{75.9} & \ACCbest{71.8} & \ACCbest{50.6} & 56.6 & \ACCbest{47.8} & \ACCbest{61.8} & \ACCbest{64.1} & \ACCbest{61.0} & 58/4.7 & \ASAbest{51/4.2} & \ASAbest{64/4.6} & 63/4.5 & 77/4.2 & \ASAbest{67/4.5} & 48/4.5 & 63/4.8 & \ASAbest{50/4.6} & \ASAbest{59/4.4} & \ASAbest{65/4.7} & \ASAbest{59/4.5} \\ 
        \midrule
        \multicolumn{25}{c}{\bf w/o images }\\
        \midrule
        Yi-Vision-V2\cite{young2024yi}  & 10.4 & 14.2 & 21.4 & 13.4 & 14.6 & 21.8 & 16.5 & 13.5 & 16.9 & 10.2 & 14.1 & 15.2 & 22/6.4 & 22/6.7 & 21/6.0 & 32/6.6 & 42/5.5 & 23/7.5 & 32/6.7 & 35/6.5 & 32/6.6 & 36/6.1 & 37/5.7 & 30/6.4 \\ 
        GPT-4o\cite{hurst2024gpt}  & 12.8 & 17.4 & 24.5 & 26.8 & 32.3 & 22.2 & 16.5 & 24.6 & 25.7 & 22.4 & 21.8 & 22.2 & 41/4.7 & 40/6.6 & 42/6.1 & 49/6.6 & 63/5.5 & 43/6.5 & 46/8.0 & 47/5.7 & 45/6.4 & 52/6.1 & 50/6.1 & 47/6.4 \\ 
        Claude-3.5-Sonnet\cite{claude-3.5-sonnet} & 22.6 & 25.5 & 34.5 & 28.4 & 40.5 & 30.1 & 22.3 & 36.5 & 29.4 & 31.5 & 23.7 & 28.8 & 29/4.8 & 38/5.2 & 47/5.0 & 48/5.3 & 69/4.7 & 43/5.2 & 44/5.0 & 46/4.9 & 44/5.1 & 50/4.6 & 46/5.0 & 45/5.0 \\ 
        Moonshot-V1-128k-Vision\cite{team2025kimi} & 25.6 & 23.9 & 32.3 & 40.2 & 43.0 & 32.4 & 27.7 & 36.5 & 27.6 & 29.5 & 28.2 & 31.2 & 37/5.2 & 29/5.7 & 41/5.1 & 40/8.5 & 55/4.7 & 43/5.1 & 38/5.1 & 45/5.4 & 40/5.3 & 46/5.0 & 48/5.3 & 41/5.6 \\ 
        GPT-4.1\cite{gpt-4.1}  & 15.2 & 23.5 & 34.5 & 39.6 & 40.5 & 40.7 & 26.9 & 35.7 & 36.0 & 31.5 & 26.3 & 31.7 & 40/6.7 & 47/8.2 & 51/7.3 & 59/7.4 & 72/6.1 & 56/7.7 & 53/6.9 & 50/6.6 & 53/7.5 & 56/6.3 & 53/7.0 & 53/7.1 \\ 
        ChatGPT-4o-Latest\cite{hurst2024gpt} & 25.0 & 29.4 & 37.7 & 37.8 & 40.5 & 38.0 & 22.8 & 32.8 & 34.2 & 43.3 & 34.3 & 33.8 & 36/6.2 & 45/8.0 & 48/7.5 & 59/7.0 & 71/5.6 & 53/7.5 & 51/6.9 & 49/6.9 & 53/8.3 & 60/6.6 & 53/6.4 & 53/7.1 \\ 
        Claude-4-Sonnet\cite{claude-4} & 32.9 & 45.8 & 52.3 & 53.7 & 63.9 & 51.9 & 37.3 & 47.1 & 37.9 & 44.5 & 46.2 & 46.2 & 31/5.7 & 53/7.0 & 55/6.1 & 66/6.4 & 84/5.5 & 61/6.2 & 55/6.3 & 54/6.2 & 56/6.4 & 62/6.3 & 50/6.2 & 57/6.3 \\ 
        Qwen-VL-Max\cite{bai2023qwen} & 37.2 & 49.0 & 51.4 & 58.8 & 72.8 & 70.8 & 44.2 & \ACCbest{55.7} & 41.9 & \ACCbest{59.4} & 51.6 & 53.2 & 40/4.9 & 45/5.0 & 55/5.0 & 63/5.0 & 82/4.3 & 63/4.8 & 51/4.7 & \ASAbest{56/5.3} & 46/5.0 & 61/4.6 & 54/5.2 & 55/4.9 \\ 
        QvQ-Max\cite{QVQ-Max} & 46.3 & 54.5 & 48.6 & 61.3 & 75.3 & 61.6 & 51.2 & 35.2 & 42.6 & 55.5 & 50.6 & 52.5 & - & - & - & - & - & - & - & - & - & - & - & - \\ 
        o4-mini\cite{o4-mini} & 28.7 & 52.9 & 53.2 & 57.6 & 65.8 & 57.9 & 53.8 & 45.5 & 57.0 & 51.2 & 41.3 & 51.6 & \ASAbest{43/4.4} & 58/4.4 & 62/4.4 & 67/4.6 & 80/4.3 & 61/4.4 & 61/4.5 & 46/4.3 & 58/4.8 & 58/4.4 & 51/4.5 & 58/4.5 \\ 
        Gemini-2.5-Flash\cite{comanici2025gemini}  & 39.0 & 68.4 & 60.5 & 66.5 & 81.6 & 69.4 & 61.3 & 48.0 & 55.9 & 54.3 & 48.4 & 59.3 & 36/5.2 & 69/5.0 & 62/4.8 & 75/5.2 & \ASAbest{91/4.5} & 76/4.7 & 67/5.1 & 51/5.3 & 59/4.6 & 63/4.5 & 58/5.1 & 64/5.0 \\ 
        Gemini-2.5-Pro\cite{comanici2025gemini} & \ACCbest{48.2} & \ACCbest{76.8} & \ACCbest{69.5} & \ACCbest{74.7} & \ACCbest{82.3} & \ACCbest{75.5} & \ACCbest{66.8} & 50.8 & \ACCbest{61.0} & 59.1 & \ACCbest{55.4} & \ACCbest{65.6} & 41/4.7 & \ASAbest{76/5.3} & \ASAbest{76/4.8} & \ASAbest{80/5.1} & 90/4.6 & \ASAbest{80/4.9} & \ASAbest{72/4.7} & 55/5.2 & \ASAbest{65/5.1} & \ASAbest{68/4.4} & \ASAbest{61/5.2} & \ASAbest{70/4.9} \\  
        \addlinespace[0.5em]
        \hdashline
        \addlinespace[0.5em]
        Mistral-Small-3.2-24B\cite{Mistral}  & 16.5 & 17.7 & 25.9 & 22.6 & 25.9 & 28.7 & 13.6 & 32.0 & 19.5 & 18.5 & 19.6 & 21.3 & 28/5.5 & 34/7.2 & 38/5.7 & 38/6.1 & 50/5.4 & 36/6.5 & 34/5.3 & 43/6.1 & 35/7.5 & 38/5.0 & 42/7.0 & 38/6.2 \\ 
        Gemma-3-27B-It\cite{team2025gemma}  &  17.7 & 19 & 23.2 & 28 & 24.1 & 27.3 & 22.5 & 27.5 & 28.3 & 18.5 & 24.4 & 23.8 & 21/7.3 & 24/9.6 & 31/8.8 & 38/8.9 & 46/7.1 & 31/8.4 & 30/7.3 & 35/8.7 & 29/8.9 & 34/6.7 & 34/8.0 & 32/8.2 \\ 
        InternVL3-14B\cite{zhu2025internvl3} & 26.2 & 32.6 & 32.7 & 38.7 & 41.8 & 44.4 & 24.3 & 43.9 & 28.3 & 30.7 & 38.5 & 34.4 & 42/5.2 & 37/5.7 & 42/5.3 & 50/5.3 & 55/4.3 & 50/4.8 & 38/4.9 & 53/5.5 & 44/5.3 & 45/4.4 & 45/4.7 & 45/5.1 \\ 
        Llama-4-Scout-17B\cite{Meta}  & 27.4 & 36.1 & 39.1 & 36.0 & 42.4 & 46.3 & 32.1 & 48.0 & 37.5 & 33.9 & 42.9 & 38.2 & 32/5.3 & 39/5.3 & 43/5.3 & 48/5.5 & 59/4.9 & 48/4.9 & 40/5.1 & 47/5.8 & 41/5.4 & 46/4.9 & 52/5.4 & 45/5.3 \\ 
        Qwen2.5-VL-32B\cite{qwen2.5} & 26.2 & 31.9 & 43.2 & 44.8 & 62.0 & 55.1 & 28.3 & 49.6 & 33.5 & 33.9 & 41.0 & 39.8 & 30/3.4 & 34/3.3 & 42/3.3 & 43/3.6 & 54/3.6 & 47/3.7 & 37/3.4 & 42/3.1 & 30/3.1 & 43/3.6 & 43/3.6 & 40/3.4 \\ 
        GLM-4.1V-9B-Thinking\cite{hong2025glm} & 33.5 & 33.9 & 46.8 & 50.6 & 60.8 & 56.0 & 40.2 & 41.0 & 39.7 & 33.1 & 44.2 & 43.0 & 49/5.0 & 39/5.0 & 55/4.7 & 61/4.8 & 74/4.3 & 62/4.8 & 53/4.7 & 55/4.9 & 46/4.6 & 48/4.4 & 57/4.9 & 54/4.7 \\ 
        Qwen2.5-VL-72B\cite{qwen2.5} & 36.6 & 44.5 & 52.7 & 52.7 & 72.8 & 56.0 & 35.3 & 54.1 & 41.5 & 44.9 & \ACCbest{51.3} & 48.3 & 38/4.8 & 46/4.8 & 58/4.7 & 60/5.3 & 78/4.4 & 60/4.7 & 49/4.4 & 56/4.9 & 43/4.7 & 54/4.3 & 60/4.6 & 54/4.7 \\ 
        Llama-4-Maverick-17B\cite{Meta} & \ACCbest{39.6} & \ACCbest{64.5} & \ACCbest{64.1} & \ACCbest{66.5} & \ACCbest{75.3} & \ACCbest{71.8} & \ACCbest{48.8} & \ACCbest{61.9} & \ACCbest{54.4} & \ACCbest{56.3} & 50.3 & \ACCbest{59.0} & \ASAbest{56/4.5} & \ASAbest{68/4.3} & \ASAbest{72/4.4} & \ASAbest{69/4.8} & \ASAbest{86/4.3} & \ASAbest{74/4.3} & \ASAbest{62/4.6} & \ASAbest{64/4.9} & \ASAbest{61/4.3} & \ASAbest{60/4.2} & \ASAbest{67/5.1} & \ASAbest{67/4.5} \\ 
        \bottomrule
    \end{tabular}
    \label{tab:complete_restructured}
\end{table*}

\textbf{Differences in the characteristics of various physics subject knowledge and VI affect model reasoning.} For mechanics problems (A-F), VI can directly provide key details such as object position and velocity direction, enabling the model to perform force analysis and dynamic calculations more directly. For electromagnetism problems (G-K), VI is equally important, but the model's textual understanding and formula application capabilities also play a large role. In these subjects, we still observe significant improvements with VI, indicating that even relatively abstract physical concepts can be better understood and reasoned by the model through concrete visual representations. Furthermore, in the evaluations with image input, models also exhibit varying capabilities across different physics subjects, which highlights the necessity of a detailed assessment of physics knowledge.

\textbf{The reasoning ability in the overall subject is deeply correlated with VI.} When image input is provided, the ACC and ASA of all MLLMs in the overall subject are significantly higher than when no image input is provided. This suggests that for complex multimodal physics problems, humans might be able to ``mentally visualize'' the described scenario to perform diagrammatic calculations, but models relying solely on the textual information in the problem statement are still insufficient to solve them. VI can significantly enhance a model's understanding and reasoning capabilities, and it has a certain universality, impacting even relatively weaker models.

\textbf{The impact of VI on the physics reasoning ability of different MLLMs varies.} Some models (e.g., o4-mini \& Gemini-2.5-Pro) show particularly significant improvements when image input is provided. These models likely possess stronger capabilities in multimodal understanding and cross-modal information fusion. They are able to more effectively extract key physical visual features from images and combine them with textual information, thereby constructing a more accurate CoT for physical reasoning to solve problems.

\textbf{Models exhibit differences under CoT evaluation.} The ASC generated by models vary. For example, in the ``w/ images'' mode, GPT-4o has an ASA of 5.3, while Gemini-2.5-Pro reaches 5.0. However, more steps are not necessarily better; some models might introduce errors in superfluous steps, thereby reducing the ASA. Furthermore, CoT accuracy is clearly correlated with the final answer accuracy; instances of low ASA but unexpectedly high ACC (e.g., Fig. \ref{fig:example}) suggest that the model might be guessing the correct answer.

\textbf{The model's CoT reasoning and instruction following have limitations.} We observe that some deep thinking models (e.g., QvQ-Max), in the ``w/o images'' mode, may not adequately follow the template for CoT evaluation, thereby affecting the implementation of CoT reasoning. This indicates VI plays a guiding role in those models' CoT generation, helping MLLMs generate reasoning steps that are more consistent with physical laws according to the context of physics problems. Meanwhile, this also illustrates that instruction following still faces complex challenges in multimodal tasks.

\begin{figure}[!t]
    \centering
    \includegraphics[width=1\linewidth]{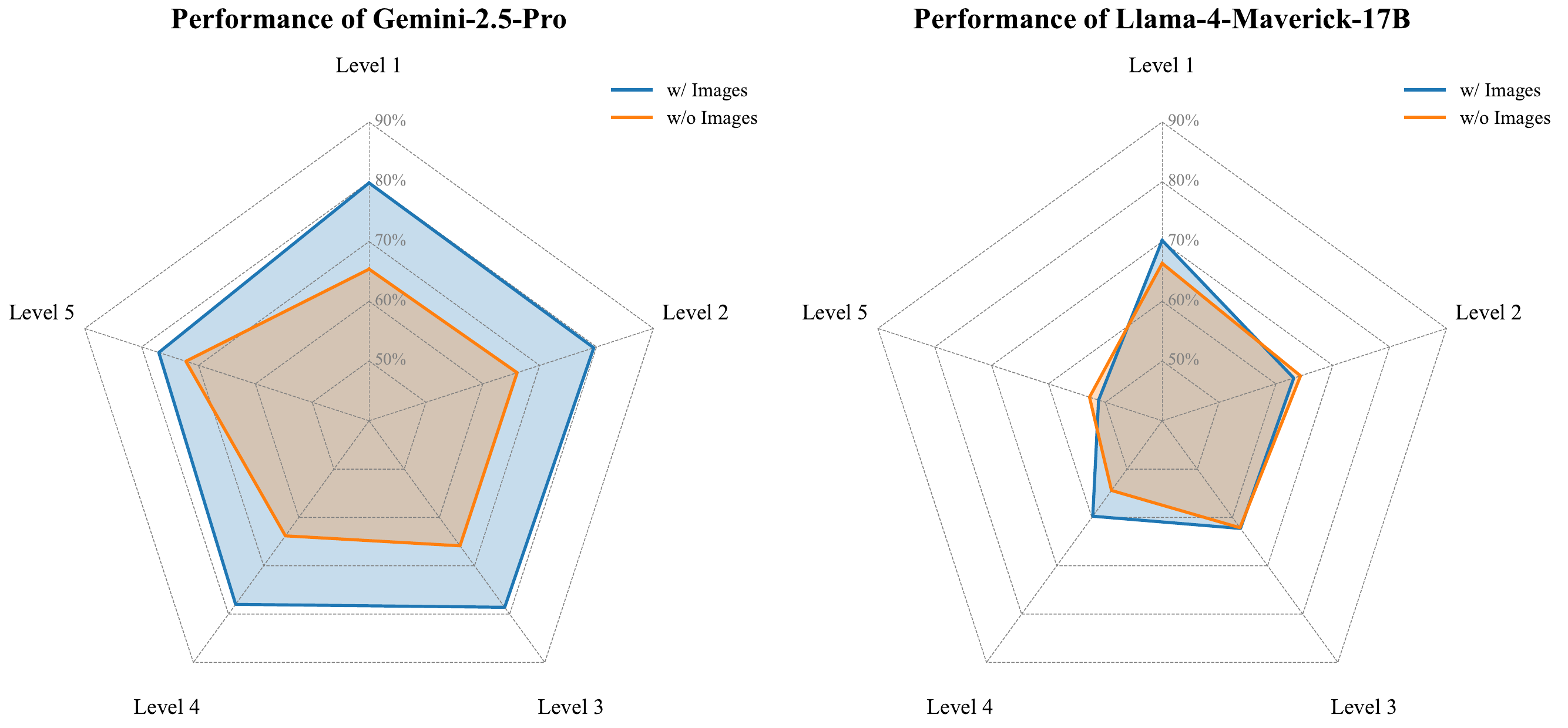}
    \caption{Performance results of the two models.}
    \label{fig:difficulty}
\end{figure}

\textbf{The influence of difficulty level and VI on different models solving problems of different difficulty levels varies.} We present the performance results of the two models, Gemini-2.5-Pro and Llama-4-Maverick-17B, as shown in Fig. \ref{fig:difficulty}. Clearly, Gemini demonstrates a relatively balanced problem-solving ability for all levels, but is significantly influenced by VI. In contrast, Llama's ability on high-difficulty problems is significantly lower than that on low-difficulty problems, but the impact of VI on it is not obvious.

\section{Conclusion}
\label{sec:Con}

We propose the Multi-Physics benchmark that includes multiple difficulty levels and utilize a detailed evaluation framework for evaluating MLLMs. This work provides rich and diverse data and insights for understanding the strengths and limitations of MLLMs when tackling Chinese multi-disciplinary physics problems of different difficulties.

\vfill\pagebreak


\bibliographystyle{IEEEbib}
\bibliography{icassp}

\vfill\pagebreak
\section{Acknowledgement}

This work was helped by volunteers, and we would like to thank them for their hard work. (Qizhi Zheng, Yi Xiao, Junyu Pan, Zhan Shen, Junhao Wu, Ya Gao, Yang Yu, Yuxi Sun, Mingxin Song, Yanzhe Fan, Peng Yang, Shuangtong Zhu, Zhongyang Cao, Qiwei Song, Mingqi Shao, Jiaming Tian, and Yuting Song)

\vfill\pagebreak

© 20XX IEEE. Personal use of this material is permitted. Permission from IEEE must be obtained for all other uses, in any current or future media, including reprinting/republishing this material for advertising or promotional purposes, creating new collective works, for resale or redistribution to servers or lists, or reuse of any copyrighted component of this work in other works.

\end{document}